\documentclass[10pt,twocolumn,letterpaper]{article}

\usepackage[pagenumbers]{cvpr} 

\usepackage[dvipsnames]{xcolor}

\usepackage{booktabs}
\usepackage{multirow}
\usepackage[accsupp]{axessibility}

\usepackage{listings}
\usepackage{ascmac}
\definecolor{cvprblue}{rgb}{0.21,0.49,0.74}
\usepackage[pagebackref,breaklinks,colorlinks,citecolor=cvprblue]{hyperref}

\def\paperID{5} 
\def\confName{GDUG}
\def\confYear{2024}

\title{SVGEditBench: A Benchmark Dataset for Quantitative Assessment of LLM's SVG Editing Capabilities}

\author{Kunato Nishina \qquad\qquad Yusuke Matsui \vspace{1mm} \\
The University of Tokyo\\
{\tt\small \{nishina, matsui\}@hal.t.u-tokyo.ac.jp}
}

\author{Kunato Nishina \qquad\qquad Yusuke Matsui \vspace{1mm} \\
The University of Tokyo\\
{\tt\small \{nishina, matsui\}@hal.t.u-tokyo.ac.jp}
}

\begin{document}
\maketitle
\begin{abstract}
Text-to-image models have shown progress in recent years. Along with this progress, generating vector graphics from text has also advanced. SVG is a popular format for vector graphics, and SVG represents a scene with XML text. Therefore, Large Language Models can directly process SVG code. Taking this into account, we focused on editing SVG with LLMs. For quantitative evaluation of LLMs' ability to edit SVG, we propose SVGEditBench. SVGEditBench is a benchmark for assessing the LLMs' ability to edit SVG code. We also show the GPT-4 and GPT-3.5 results when evaluated on the proposed benchmark. In the experiments, GPT-4 showed superior performance to GPT-3.5 both quantitatively and qualitatively. The dataset is available at \url{https://github.com/mti-lab/SVGEditBench}.
\end{abstract}
\vspace{-10pt}    
\section{Introduction}
\label{sec:intro}

Vector graphics are popular for various applications because of the features not found in raster images. Vector graphics uses primitive shape elements such as circles and squares to represent a scene. Since vector representation expresses each element in the scene individually, they are highly editable~\cite{svgdreamer}. Also, one of the most prominent features of vector graphics is that the image quality will not degrade when displayed in any size. Scalable Vector Graphics (SVG)~\cite{SVG} is the representative vector graphics format used as a standard in web icons and fonts.

With the recent advancements in Large Language Models (LLMs), generating and editing vector graphics is now possible with LLMs. Research shows that LLMs like ChatGPT~\cite{chatgpt} and GPT4~\cite{gpt4} can perform various tasks. Those tasks include generating programming code and summarizing or translating text ~\cite{gpt3, bubeck2023sparks}. Since an SVG file is not a binary but a text file (XML), LLMs can directly handle those files. Hence, we could use LLMs to process SVG.
Image generation models with diffusion have advanced in recent years~\cite{dalle3, stable-diffusion, diffusion}, but combining such models with LLMs is still challenging. SVG processing with LLMs means we do not have to use those generation models. Also, using communicative LLMs such as ChatGPT~\cite{chatgpt}, editing vector graphics can be realized through text chat. Since vector graphics editing typically requires knowledge and specialized software~\cite{iconshop}, being able to use intuitive interfaces like text chat can be a great advantage.

Research on SVG generation or editing with LLMs exists~\cite{bubeck2023sparks, cai2023leveraging}. However, they only provide examples and do not quantitatively show how LLMs can handle the numerous SVG editing tasks.

In this paper, we built a benchmark dataset that quantitatively evaluates LLMs' SVG editing capabilities. We selected six editing tasks whose quality can be measured easily. We also created the LLM prompt and the model response for each editing task. Comparisons of the capabilities between models will be possible with this benchmark. Additionally, we conducted experiments on GPT-4 and GPT-3.5 with the proposed benchmark. We examined its validity by comparing the results with qualitative evaluations. GPT-4 outperformed GPT-3.5 in all six editing tasks. Both quantitative and qualitative experiments confirmed this tendency.
\section{Related Works}

\subsection{Scalable Vector Graphics}
An example of an SVG code and its rendered result is shown in Figure \ref{fig:svg}. SVG uses XML format that takes an \texttt{<svg>} tag as its top-level element to represent a scene. The root \texttt{<svg>} tag contains tags representing shapes or text as its child. Those tags fall into three main categories: basic shapes such as rectangles (\texttt{<rect>}) and circles (\texttt{<circle>}), curves composed of straight lines and B\'ezier curves (\texttt{<path>}), and text (\texttt{<text>}). Each tag has its own set of attributes (e.g., \texttt{cx}, \texttt{fill}, \texttt{d} in Figure \ref{fig:svg}) that define the position or color of the shape. Since paths can also express basic shapes, using paths is more flexible. This expressivity of paths is why most previous works in the next section learn models that only deal with paths. However, we cannot quickly determine the shape by looking at the path representation, especially when the shape is complex. This is because paths are represented only by combinations of the line type and the coordinates of their control points. The \texttt{<path>} elements also show no semantic information.

{
\setlength\textfloatsep{5pt}
\begin{figure}
    \centering
    \includegraphics[width=\hsize]{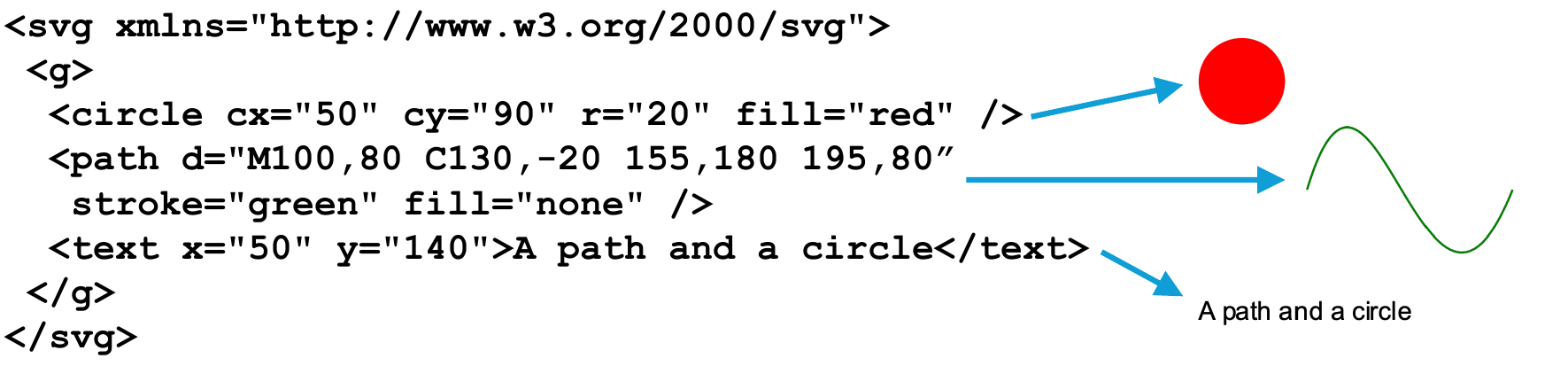}
    \caption{An example of an image represented in SVG format. Each XML element corresponds to a single shape or text block, as indicated by the blue arrows.}
    \label{fig:svg}
\end{figure}
}

\subsection{Recent Studies on Vector Image Processing}
Regarding research related to vector graphics, especially SVG, various processing tasks and models have been proposed within the last few years. Popular tasks include:
\begin{itemize}
    \item Vectorization: converts a raster image into a vector image
    \item Text-to-vector: generating a vector image conditioned by an input text
    \item Editing: edits the input vector image in a specific way (the focus of this paper)
\end{itemize}

Vectorization methods have advanced using the rasterization methods, especially DiffVG~\cite{DiffVG}. Im2Vec~\cite{Im2Vec} uses RNN for vectorization. LIVE~\cite{LIVE} proposed a method to progressively add the number of shapes to represent the scene. A recent method, S\textsuperscript{2}VG\textsuperscript{2}~\cite{S2VG2}, shows that a combination of Vision Transformer~\cite{vision-transformer} and language models (BERT~\cite{bert}) can generate human-readable SVG code.

Text-to-vector is a significant research topic in recent vector graphics processing. Its methods can be broadly classified by whether or not they use diffusion models~\cite{diffusion}. Examples not using diffusion include IconShop~\cite{iconshop} and StrokeNUWA~\cite{strokenuwa}. IconShop generates icons using an Autoregressive Transformer~\cite{transformer}. StrokeNUWA learns tokens representing strokes, and an LLM uses those tokens to generate a vector image. On the other hand, research that uses diffusion models includes VectorFusion~\cite{vectorfusion} and SVGDreamer~\cite{svgdreamer}. They integrate DiffVG and diffusion models in a loop that optimizes the SVG parameters.

Concerning editing, Zhang et al.~\cite{svg-customization} attempt to customize vector images via a text prompt. DiffVG first renders the input image into a raster. Then, a diffusion model edits the rendered image, and SVG paths are optimized while semantically aligning with the edited raster image.

Some examples try to perform the SVG image processing tasks mentioned above with LLMs. For vectorization, StarVector~\cite{starvector} outputs SVG code with an LLM for code, named  StarCoder~\cite{starcoder}. Several works~\cite{bubeck2023sparks, cai2023leveraging} show chat-based SVG editing and generation examples. SVG manipulation using LLMs is a research area that is gaining momentum.
\section{Building the Benchmark}
This section provides the details of the editing task used for the benchmark. We show the method we used to select the original SVG image, the details of the six editing tasks, and the quantitative evaluation method.

\subsection{Overview of the Tasks}
\begin{figure*}
    \centering
    \includegraphics[width=0.75\hsize]{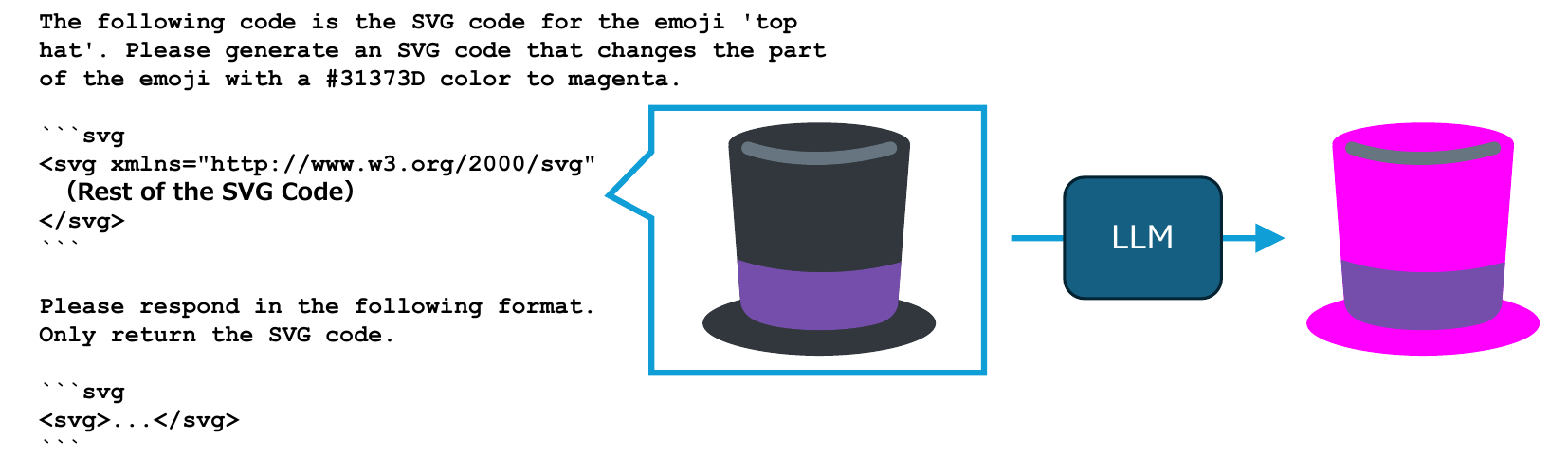}
    \caption{An overview of the tasks in the proposed benchmark and an example of the prompt in the \textbf{Change Color} task.}
    \label{fig:task-example}
\end{figure*}
Figure \ref{fig:task-example} shows an overview of the task used in the proposed benchmark. The prompt given to the evaluated LLM consists of the following three parts. Firstly, it explains the editing task the LLM should perform. Secondly, it provides the SVG code before the edit. Finally, it specifies the format in which LLM should respond. We regard the text between \texttt{```svg} and \texttt{```} as the output image. We render the output SVG code into PNG before evaluating the editing quality numerically. For some tasks, we also use the code itself for evaluation. Refer to Section \ref{sec:tasks} for more details.

\subsection{Selection of SVG Data}\label{sec:dataset}
We selected Twemoji~\cite{twemoji} as the SVG data before editing in the proposed benchmark. This decision was under the following criteria. Firstly, the data should be easily retrievable as SVG files. Secondly, the SVG images should contain both \texttt{<path>} elements and other primitive shape elements. Thirdly, the SVG file should be small, and lastly, an explanation text for each image should be available. Twemoji~\cite{twemoji} is part of the benchmark used in SVGBench. SVGBench is a method of evaluating SVG generation models proposed with StarVector~\cite{starvector}. Twemoji contains 3689 pairs of SVG code and $72\times72$ PNG image of emojis corresponding to Unicode 14.0~\cite{unicode14}.

\begin{figure}[t]
    \centering
    \begin{tabular}{c}
         \includegraphics[width=0.15\hsize]{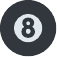}  
         \includegraphics[width=0.15\hsize]{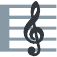}  
         \includegraphics[width=0.15\hsize]{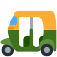}
         \includegraphics[width=0.15\hsize]{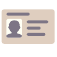}\\ \midrule
         \includegraphics[width=0.15\hsize]{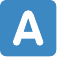}  
         \includegraphics[width=0.15\hsize]{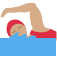}
         \includegraphics[width=0.15\hsize]{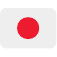}  
         \includegraphics[width=0.15\hsize]{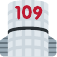}
    \end{tabular}
    \caption{Sample images in the Twemoji dataset. The top row shows some images in the dataset, and the bottom row shows the ones removed.}
    \label{fig:twemoji-example}
\end{figure}

We further filtered the images in Twemoji. Firstly, we removed the Regional Indicator Symbols\footnote{For instance, Regional Indicator Symbols for J and P show an emoji of the Japanese national flag.} since they contain the phrase \texttt{REGIONAL INDICATOR SYMBOL LETTER} in its name. This name is unrelated to the appearance of the emoji, and we considered that this 
could lead to confusion by the LLM. Also, we removed ZWJ sequences\footnote{Multiple Unicode characters can be joined into a single glyph with the ZERO WIDTH JOINER (U+200D). These sequences of characters are called ZWJ sequences.}~\cite{zwjsequence} and flags. We also removed the emojis whose names are unavailable with the \texttt{unicodedata} library in Python 3.12. The above process resulted in 1366 images. Figure~\ref{fig:twemoji-example} shows some examples of images in the benchmark and some removed images.

\subsection{Evaluation Tasks and Metrics} \label{sec:tasks}
\begin{figure}[t]
    \centering
    \footnotesize
    \begin{tabular}{c|ccc}
    Original & \textbf{Change Color} & \textbf{Set Contour} & \textbf{Compression} \\
    \multirow{3}{*}{\includegraphics[width=0.15\hsize]{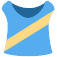}} & \includegraphics[width=0.15\hsize]{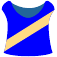} & \includegraphics[width=0.15\hsize]{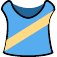} & \includegraphics[width=0.15\hsize]{img/Fig5-1.pdf} \\
    &\textbf{Upside-Down} & \textbf{Transparency} & \textbf{Crop to Half} \\
    & \includegraphics[width=0.15\hsize]{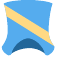} & \includegraphics[width=0.15\hsize]{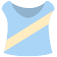} & \includegraphics[width=0.075\hsize]{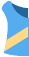} \\
    \end{tabular}
    \caption{Examples of answers for each task used in the proposed benchmark. Note that for the \textbf{Compression} task, the rendered result should not change from the original.}
    \label{fig:answers}
\end{figure}
We created the LLM prompts for performing the following six tasks. We also generated the images after correct modification (answers) using the emoji SVG data obtained above. We selected these six tasks by considering whether we could generate the answers automatically. As shown below, most tasks here can be achieved only by changing a single attribute of the SVG code. Therefore, these tasks can test the LLMs if they know the SVG functionality. Figure~\ref{fig:answers} shows the answers for each task. We explain the structure of the prompts in more detail in the supplementary material.

\begin{description}
    \item[Change Color] We randomly selected a color from the ones specified in the \texttt{fill} attribute. The task is to change the color of the part with the picked color into another. We picked the target color randomly from red, green, blue, yellow, cyan, magenta, white, and black. Modifying the applicable \texttt{fill} attribute is adequate for this change.
    \item[Set Contour] We selected a color from the image, similarly to the \textbf{Change Color} task. The task is to draw a black line around the part with the picked color. Setting the \texttt{stroke} and the \texttt{stroke-width} attributes achieves this modification.
    \item[Compression] In this task, we asked the LLM to shorten the SVG code without changing the appearance. For instance, replacing the shapes bounded by four straight lines to \texttt{<rect>} or \texttt{<polygon>} may make the SVG expression shorter. The LLM has to look at the input code throughout, interpret its graphical meaning, and look for parts of the code to compress. Therefore, this task is more complex than others.
    \item[Upside-Down] The task is to flip the image upside down. Adding an appropriate \texttt{transform} attribute to the root \texttt{<svg>} element can suffice the task.
    \item[Transparency] The task is to make the image half transparent. Setting the \texttt{opacity} attribute of the root element to \texttt{0.5} can perform the task.
    \item[Crop to Half] The task is to trim the right half of the input image and only leave the left half. This modification can be accomplished by editing the \texttt{viewBox} attribute of the \texttt{<svg>} and setting the width to half.
\end{description}

Except for the \textbf{Compression} task, we generated the answer by replacing or adding the attributes shown above to the SVG code in Twemoji. We converted the LLM output and the answer SVG into a $72\times72$ PNG using the CairoSVG library~\cite{cairosvg}. We compare the two converted images by calculating the Mean Squared Error (MSE) between the two raster images. Since the MSE does not consider the SVG code, the output is evaluated as correct if the edits by the LLMs are equivalent. For example, replacing the shapes with the ones with half the width may accomplish the \textbf{Crop to Half} task.

We set the background to white when converting to PNG. This setting ensures correct comparison with MSE for the transparent areas. Also, we standardized the pixel values to fall between 0 and 1. Since we averaged the MSE calculated for each color channel, the MSE for a single image will also be between 0 and 1.

The answer to the \textbf{Compression} task is the SVG code in Twemoji, and the MSE was calculated similarly to the other tasks. In addition to MSE, we calculated the compression rate by comparing the length of the SVG code ((output code length) / (input code length)). 

We randomly selected one hundred emojis from the 1366 emojis selected in the previous section. Prompts and answers for the six tasks were created for each emoji and used as the evaluation dataset. Therefore, a single LLM performed 600 SVG edits in total.
\section{Experiments}
\begin{table}[t]
    \centering
    \caption{Results of evaluating GPT-4 and GPT-3.5 with the proposed benchmark. We also show the results when no edits are made as a reference.}
    \small
    \begin{tabular}{@{}lllll@{}}
        \toprule
        & & \multicolumn{2}{c}{Model} &  \\ \cmidrule(lr){3-4}
        Task & Metric↓  & GPT-4 & GPT-3.5 & No Edit \\ \midrule
        \textbf{Change Color} & MSE & $\mathbf{6.88\times 10^{-5}}$ & 0.0134 & 0.0702 \\
        \textbf{Set Contour} & MSE & \textbf{0.0190} & 0.0362 & 0.0286 \\
        \multirow{2}{*}{\textbf{Compression}} & Ratio  & \textbf{94.5\%} & 96.1\% & 100\%\\
                               & MSE &  0.0071 & \textbf{0.0023} & 0 \\
        \textbf{Upside-Down} & MSE & \textbf{0.0463}  &  0.0705 & 0.0878 \\
        \textbf{Transparency} & MSE & \textbf{0.0012} & 0.0122 & 0.0402 \\
        \textbf{Crop to Half} & MSE & \textbf{0.0851} & 0.1068 & 0.1174 \\
        \bottomrule
    \end{tabular}
    \label{tab:result}
\end{table}

This section presents the results of evaluating the SVG editing capabilities of GPT-4/3.5 with the proposed benchmark.
\subsection{Quantitative Evaluation of GPT Models}
We sent the prompts in the dataset to GPT-4 and GPT-3.5 via the OpenAI API~\cite{openai-api}. The models used were \texttt{gpt-4-0125-preview} and \texttt{gpt-3.5-turbo-0125}. To ensure reproducibility, we set the temperature parameter to \texttt{0}. We calculated the MSE and the compression ratio using the SVG obtained from the model outputs. Then, we averaged the values over the 100 images for each task. We did not include the result in the average calculation if there were no or multiple valid SVG codes in the response. Table~\ref{tab:result} shows the evaluation results. We also show the results without edits to the images in the table's rightmost column. If the metrics are smaller than this value, it indicates that the LLMs could perform the task.

The results reveal that GPT-4 outperformed GPT-3.5 in terms of MSE except for the \textbf{Compression} task. Even for the \textbf{Compression} task, the compression ratio is smaller with GPT-4, which means GPT-4 has higher compression capabilities. Therefore, GPT-4 shows better editing performance with the quantitative evaluation by the proposed method.

\subsection{Comparison with the Qualitative Evaluation}
\begin{figure}[t]
    \centering
    \begin{tabular}{cccc}
        Original & Answer & GPT-4 & GPT-3.5 \\
        \includegraphics[width=4em]{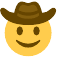} & \includegraphics[width=4em]{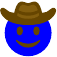} & \includegraphics[width=4em]{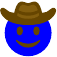} & \includegraphics[width=4em]{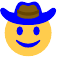}\\ \midrule
        \includegraphics[width=4em]{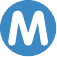} & \includegraphics[width=4em]{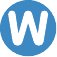} & \includegraphics[width=4em]{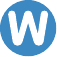} & \includegraphics[width=4em]{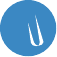}\\
        
    \end{tabular}
    \caption{A qualitative evaluation of the SVG editing results generated by LLMs. The two rows show an example of the \textbf{Change Color} task and the \textbf{Upside-Down} task, respectively.}
    \label{fig:qualitative}
\end{figure}

In this section, we discuss the factors that led to the difference in performance between the two models by looking into the actual output image. We point out two differences. The first point is that GPT-4 could reflect the instructions to the output more appropriately. The upper row of Figure~\ref{fig:qualitative} is one example. GPT-3.5 painted the parts with a color different from the one indicated in the prompt. The second point is that GPT-3.5 often redrew the paths unnecessarily. Images in the bottom row of Figure~\ref{fig:qualitative} are one example from the \textbf{Upside-Down} task. As mentioned in Section \ref{sec:tasks}, adding or editing one attribute can complete the tasks used here, except for the \textbf{Compression} task. However, GPT-3.5 rewrote the paths, which resulted in significant image corruption.

Also, for the \textbf{Compression} task, we found that GPT-4 rounded off numbers in the coordinates to shorten the representation. Although the prompts did not suggest that rounding leads to shorter code, it is interesting that GPT-4 recognized that this strategy would compress the representation while maintaining the image nearly unchanged.

The above qualitative results show that GPT-4 is superior to GPT-3.5 in SVG editing capabilities. This result is consistent with our quantitative experiments. We conclude that the metrics in the proposed benchmark reflect the LLMs' editing performance.
\section{Conclusion}
Considering the abilities of LLMs in handling SVG editing, we built a benchmark dataset to evaluate the SVG editing performance of LLMs quantitatively. 
We evaluated GPT-4 and GPT-3.5 with the proposed benchmark and showed that GPT-4 outperforms GPT-3.5 in SVG editing. This trend was also true when we compared the actual output images.

Future directions would be to improve the dataset by adding tasks, especially the ones that test semantic understanding of SVGs. Also, the benchmark can be used not only for existing models like GPT-4 and GPT-3.5 but for LLMs finetuned for SVG editing.
{
    \small
    \bibliographystyle{ieeenat_fullname}
    \bibliography{main}
}

\clearpage
\setcounter{page}{1}
\renewcommand{\maketitlesupplementary}
{
\newpage
\onecolumn
    {
        \centering
        \Large
        \textbf{\thetitle}\\
        \vspace{0.5em}Supplementary Material \\
        \vspace{1.0em}
    }
}
\newcommand{\colored}[2]{{\bfseries\color{#1}#2}}

\maketitlesupplementary

\section{Structure of the Prompts}
\label{sec:supplementary}

Here, we show how we composed the prompts we input to the LLMs in the proposed benchmark dataset. The following prompt is an example of the \textbf{Change Color} task.

\begin{itembox}[l]{An example prompt of the \textbf{Change Color} task}
    \begin{lstlisting}
<r#The following code is the SVG code for the emoji#> <p#'movie camera'#><r#. Please generate an SVG code that changes the part of the emoji with a#> <g##31373D#> <r#color to#> <g#red#><r#.#>

<b#```svg
<svg xmlns="http://www.w3.org/2000/svg" viewBox="0 0 36 36"><path fill="#31373D" d="M32 21v1h-2v-1c0-.446-.09-.867-.225-1.268 2.446-.757 4.224-3.038 4.224-5.733 0-3.314-2.687-6-6-6-1.603 0-3.055.632-4.131 1.656C23.241 6.433 20.405 4 17 4c-3.866 0-7 3.134-7 7 0 2.551 1.369 4.777 3.409 6H13c-2.209 0-4 1.791-4 4H8l-6-4H1v14h1l6-4h1v2c0 2.209 1.791 4 4 4h13c2.209 0 4-1.791 4-4v-3h2v1h3v-6h-3z"/><path fill="#66757F" d="M22 11c0 2.761-2.239 5-5 5s-5-2.239-5-5 2.239-5 5-5 5 2.238 5 5z"/><circle fill="#CCD6DD" cx="17" cy="11" r="2"/><circle fill="#66757F" cx="27.999" cy="14" r="4"/><circle fill="#CCD6DD" cx="27.999" cy="14" r="2"/><path fill="#8899A6" d="M17 20h10v10H17z"/><path fill="#31373D" d="M19 22h6v6h-6z"/><circle fill="#8899A6" cx="12.999" cy="28" r="2"/></svg>
```#>

<o#Please respond in the following format. Only return the SVG code. 

```svg
<svg>...</svg>
```#>\end{lstlisting}
\end{itembox}

The \colored{red}{red part} at the beginning of the prompt describes the editing task. The description includes the emoji's name (\colored{RoyalPurple}{purple part}). We show the name to specify what the following SVG code represents. The \texttt{name} function of the \texttt{unicodedata}\footnote{\url{https://docs.python.org/ja/3/library/unicodedata.html}} Python standard library returns the name of an emoji as defined in Unicode. For the \textbf{Change Color} and the \textbf{Set Contour} tasks, we must indicate the area to edit. We randomly choose a fill color from the ones in the SVG code and show the color in the \colored{Green}{first green part} of the prompt. The LLM should edit the part filled with the chosen color. We also randomly select a target color among red (\texttt{\#FF0000}), green (\texttt{\#00FF00}), blue (\texttt{\#0000FF}), yellow (\texttt{\#FFFF00}), cyan (\texttt{\#00FFFF}), magenta (\texttt{\#FF00FF}), white (\texttt{\#FFFFFF}), and black (\texttt{\#000000}). The \colored{Green}{second green part} denotes the target color.

The \colored{Cyan}{cyan part} is the SVG code before the edit. The code is surrounded by \texttt{```svg} and \texttt{```} to explicitly indicate to the LLM that it is the code part. Finally, the \colored{Orange}{orange part} defines the output format. We also state that the LLM should only return the SVG code to supress explaining the content of the image or the procedures of editing. We can automatically process the LLM's output by adding this paragraph. Namely, we regard the text surrounded by \texttt{```svg} and \texttt{```} as the LLM's output.

\newpage
\section{Example of Prompts for Each Prompt}
This section shows the prompts used in each task except for the \textbf{Change Color} task, which we already demonstrated. These prompt examples are for the same emoji as the previous example. Note that we omitted the \colored{Cyan}{cyan part} and the \colored{Orange}{orange part} since these parts are the same as earlier.

\begin{itembox}[l]{\textbf{Set Contour}}
    \begin{lstlisting}
<r#The following code is the SVG code for the emoji#> <p#'movie camera'#><r#. Please generate an SVG code that draws a black line around the part of the emoji with a#> <g##66757F#> <r#color.#>
    \end{lstlisting}
\end{itembox}
\begin{itembox}[l]{\textbf{Compression}}
    \begin{lstlisting}
<r#The following code is the SVG code for the emoji#> <p#'movie camera'#><r#. Please generate a more compact SVG code that represents the same emoji.#>
    \end{lstlisting}
\end{itembox}
\begin{itembox}[l]{\textbf{Upside-Down}}
    \begin{lstlisting}
<r#The following code is the SVG code for the emoji#> <p#'movie camera'#><r#. Please flip this emoji upside down.#>
    \end{lstlisting}
\end{itembox}
\begin{itembox}[l]{\textbf{Transparency}}
    \begin{lstlisting}
<r#The following code is the SVG code for the emoji#> <p#'movie camera'#><r#. Please make this emoji transparent by half.#>
    \end{lstlisting}
\end{itembox}
\begin{itembox}[l]{\textbf{Crop to Half}}
    \begin{lstlisting}
<r#The following code is the SVG code for the emoji#> <p#'movie camera'#><r#. Please trim the right half and keep the left half.#>
    \end{lstlisting}
\end{itembox}

\end{document}